\documentclass[runningheads]{llncs}
\usepackage{graphicx}
\usepackage{amsmath}
\usepackage{hyperref}
\usepackage{amssymb}
\usepackage{amsfonts}
\usepackage{pifont}

\usepackage{bm} 
\usepackage{fixmath} 
\usepackage[table]{xcolor}

\begin{document}
\title{Trilateral Attention Network for Real-time Medical Image Segmentation}
%
\author{Ghada Zamzmi\inst{1} \and
Vandana Sachdev \inst{2} \and
Sameer Antani\inst{1}}
%
%
\institute{Lister Hill National Center for Biomedical Communications \and National Heart, Lung, and Blood Institute \\ U.S. National Institutes of Health (NIH), Bethesda, MD, 20892}

%
\maketitle              

\begin{abstract}
Accurate segmentation of medical images into anatomically meaningful regions is critical for the extraction of quantitative indices or biomarkers. The common pipeline for segmentation comprises regions of interest detection stage and segmentation stage, which are independent of each other and typically performed using separate deep learning networks. The performance of the segmentation stage highly relies on the extracted set of spatial features and the receptive fields. In this work, we propose an end-to-end network, called Trilateral Attention Network (TaNet), for real-time detection and segmentation in medical images. TaNet has a module for region localization, and three segmentation pathways: 1) handcrafted pathway with hand-designed convolutional kernels, 2) detail pathway with regular convolutional kernels, and 3) a global pathway to enlarge the receptive field. The first two pathways encode rich handcrafted and low-level features extracted by hand-designed and regular kernels while the global pathway encodes high-level context information. By jointly training the network for localization and segmentation using different sets of features, TaNet achieved superior performance, in terms of accuracy and speed, when evaluated on an echocardiography dataset for cardiac segmentation. The code and models will be made publicly available in \href{https://github.com/TaNet}{TaNet Github page}.

\end{abstract}
%
%

\section{Introduction}
Medical image segmentation divides the images into different semantic or anatomical regions, based on which important quantitative indices can be extracted for disease diagnosis. It has been widely applied to various medical imaging modalities including X-ray \cite{gaal2020attention}, ultrasound \cite{ouahabi2021deep}, and magnetic resonance imaging \cite{cui2018automatic}, to delineate anatomical regions for further quantitative analysis. Current medical image segmentation approaches can be divided into global-based and region-based. Global-based approaches use the entire image while the region-based approaches focus on a specific region of interest (ROI). The majority of region-based approaches use two separate networks for ROI localization stage and semantic segmentation stage. However, combining localization and segmentation into a single network is efficient as it prevents unnecessary repetitions of training separate networks in isolation \cite{zamzmi2020accelerating,zamzmi2021ums,he2017mask,cao2019triply}. Further, jointly training the network for localization and segmentation improves the performance as it optimizes segmentation based on localised ROIs; e.g., recent works \cite{gu2020net,sinha2020multi} demonstrated that the incorporation of single pathway \cite{gu2020net} and dual pathway \cite{sinha2020multi,gu2020net} attention mechanisms into segmentation networks allowed to localize the target ROI while suppressing irrelevant parts, leading to increased segmentation accuracy. 

The spatial and global (context) information are both essential to the semantic segmentation task \cite{taghanaki2021deep}. Earlier methods for spatial information extraction relied on hand-designed features generated by classical descriptors; e.g., local binary pattern \cite{iakovidis2008fuzzy,pereira2020dermoscopic}) and atlases \cite{bazin2008homeomorphic,daly2020multiatlas}. Different hand-designed methods can capture unique statistical, geometrical, or textural features from the images and are optimized for performance and power efficiency \cite{ma2021image}. In recent years, deep learning-based methods for segmentation (e.g., fully convolutional networks \cite{long2015fully}, U-Net \cite{ronneberger2015u}) have shown superior performance as they capture rich low-level details at different levels of abstraction. These deep learning methods can be divided, based on their architectures, into: dilation architecture and encoder-decoder architecture. The dilation architecture uses dilated convolutions to preserve high-resolution feature representations. Examples of current state-of-the-art architectures with dilated convolutions include DeepLabv3 \cite{yurtkulu2019semantic} and PSPNet \cite{weng2019automatic}. The encoder-decoder architecture has downsampling and upsampling components and uses skip connections to capture wider context and recover the high-resolution feature representations. Examples of well-known encoder-decoder architectures include U-Net \cite{ronneberger2015u} and SegNet \cite{badrinarayanan2017segnet}. 

Although these architectures achieved excellent segmentation performance, the dilation convolutions and skip connections increase the computational complexity and memory overhead, leading to a slow inference speed. To solve this issue, Yu et al. \cite{yu2020bisenet} proposed a two-pathway architecture, known as the Bilateral Segmentation Network (BiSeNet), to speed up the inference time. The first pathway has wide channels and shallow layers to capture the spatial information while the second pathway provides a large receptive field to capture wider context. The two pathways concurrently generate the feature representations, which significantly increases the efficiency. As demonstrated in \cite{yu2018bisenet}, this conceptual design is significantly faster than both dilation and encoder-decoder architectures. 

In this work, we extend BiSeNet and propose an efficient end-to-end Trilateral Attention Network (TaNet) for real-time medical image localization and segmentation. Our specific contributions are summarized as follows. First, we integrate into BiSeNet a third handcrafted pathway ($HP$) with handcrafted-encoded convolutional kernels. This pathway can be used to extract unique set of morphological, statistical, or textural attributes. Second, we integrate a Spatial Transformer Network (STN) as a module to direct the attention of the segmentation task to specific ROIs while learning the context relationship between them. Third, we present a strategy to jointly train the entire network (localization + segmentation) for ROI-based semantic segmentation. TaNet achieves impressive real-time results when evaluated on the echocardiography dataset for cardiac segmentation. It outperforms BiSeNet \cite{yu2018bisenet}, FCN \cite{long2015fully}, U-Net \cite{ronneberger2015u}, and recent methods for echocardiography segmentation \cite{ouyang2020video,ali2021echocardiographic}.

\section{Echocardiogram Dataset}
We used XXX echocardiogram dataset, which contains 300 videos (parasternal long axis view or PLAX) recorded from 300 patients who were referred for echocardiographic examination in the Clinical Center at XXX. The segmentation ground truth (GT) masks, which were provided by experts, contain semantic labels for the following cardiac regions: left ventricle (LV), left atrium (LA), right ventricular outflow tract (RV), interventricular septum (IVS), and posterior wall (PW). Examples from our dataset are provided in the supplementary materials. In this work, we divided (subject-wise) the entire dataset into training (80\%) and testing set (20\%). The training set is further divided into training and validation using 10-fold cross validation. To enlarge the training set, we applied the following operations: random rotation ($-15^{\circ}$ to $+15^{\circ}$), horizontal and vertical shift ($-0.25$,$0.25$), scale $[0.75, 1, 1.25]$, and horizontal and vertical flip. We resized (bicubic interpolation, OpenCV-python) all the images into $512 \times 512$.



\section{Trilateral Attention Network (TaNet)}
Figure 1 depicts our proposed network for ROI-based medical image segmentation. TaNet integrates localization into the segmentation by adding the STN module and uses three pathways for segmentation. The entire network is trained end-to-end using a joint learning approach to prevent unnecessary training repetitions, and allows the network to focus on specific ROIs (i.e., cardiac regions) while learning the context relationships among them. In the traditional approach, localization and segmentation are trained separately, which adds unnecessary training repetitions and disregards the relationship between both stages. 

   \begin{figure}[!t]
        \centering
        \includegraphics[height=0.45\textwidth]{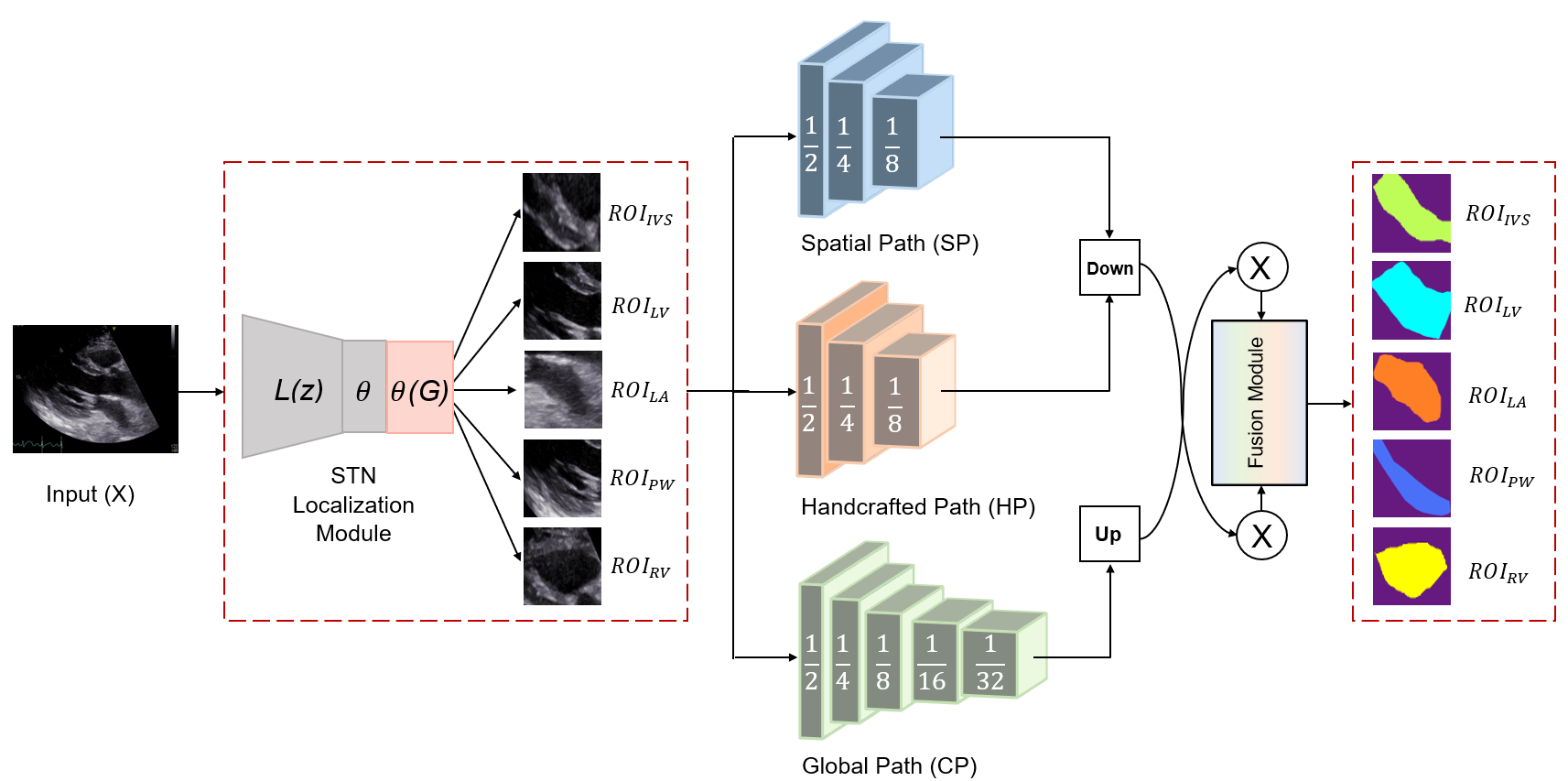}
        \caption{Overview of TaNet with its components: STN for ROIs localization and segmentation backbone with 3 pathways, spatial (detail) path ($SP$), handcrafted path ($HP$), global path ($GP$). The numbers in the cubes are the size ratios to the resolution of the input. In the fusion module, we aggregate the feature maps from all pathways. Down, up, and $\otimes$ indicate downsampling, upsampling, and element-wise product, respectively. The localization module focuses the segmentation attention to different ROIs ($X$). The output of the network is an image with the segmented ROIs  and the background.}
        \label{fig:my_label}
    \end{figure}

\subsection{ROIs Localization}
STN \cite{jaderberg2015spatial} is a differentiable module that can be inserted anywhere in convolutional networks and used for applying spatial transformations to the input image or feature map. It is differentiable in the sense that it computes the derivative of the transformations in the module allowing to learn the gradients of the loss with respect to the module parameters. This allows the training of STN with any convolutional network end-to-end. STN consists of a localization network ($L$), a grid generator ($G$), and a sampler. The localization network learns the spatial transformation parameters ($\theta$). The grid generator creates, based on $\theta$, the coordinates where the input map should be sampled to obtain the transformed map. Finally, the sampler produces the desired transformed map.

In medical images, it is common that the target ROI occupies a relatively small portion of the image \cite{hesamian2019deep,khor2017region,badshah2020new}. Hence, considering the entire image for segmentation would add noise caused by irrelevant portions in the background and lead to the segmentation network being biased toward the background. In this work, we integrate STN and use it as a method for focusing the attention of the segmentation on a specific ROI while suppressing the potential irrelevant portions and learning the context relationship among different ROIs. \textit{We hypothesize that the integration of STN would improve the segmentation performance.}

\subsubsection{Localization Network}
Prior to the use of localization network, we use FCN-8 \cite{long2015fully,yin2020end} model for coarse segmentation. Providing a coarse segmentation of different ROIs allows the localization network to 1) generate the transformation parameters ($\theta$) for these regions and 2) learn the context relationship among them. The output of FCN-8 is an image with coarse semantic labels corresponding to the cardiac regions (ROIs). Given an input image $X \in \mathbb{R}^{C \times H \times W}$, where $C$, $H$, and $W$ represent the channels, width, and height, respectively, the output of the coarse segmentation can be expressed as:

\begin{equation} 
    z = FCN (X)
\end{equation}

where $z \in \mathbb{R}^{H \times W}$. The coarse mask ($z$) is then sent to the localization network ($L$) to generate the spatial transformation matrices ($\theta$) for $N$ cardiac regions (ROIs):

\begin{equation} 
\theta = L(z)
\end{equation}

where $\theta \in \mathbb{R}^{N \times 2 \times 3}$; here $N=5$ as  there are five cardiac regions as shown in Figure 1. As for the localization network ($L$), we used a simplified version of VGG16 \cite{yin2020end} that has 8 convolutional layers and a final regression layer to generate $N \times 2 \times 3$ spatial transformation matrix ($\theta$). $L$ outputs the spatial transformation matrix ($\theta$) as shown in Figure 1. For each ROI, $\theta$ is defined as follows:

\begin{equation}
             \theta = \begin{bmatrix}
              s_x & 0 & t_x\\
             0 & s_y & t_y
             \end{bmatrix}
\end{equation}

where $s_x$, $s_y$, $t_x$, and $t_y$ parameters, which are learned by $L$, allows ROIs scaling and translation. 

\subsubsection{Grid Generator \& Bilinear Sampler} Given $\theta \in \mathbb{R}^{N \times 2 \times 3}$, the relevant parts of the image (i.e., $ROI_i$, $i \in \{1,2,..,N\}$) are sampled into a sampling grid $G$ of pixels $G_i =  (x_i^t, y_i^t)$ to form an output feature map $V \in \mathbb{R}^{C \times H' \times W'}$, where $C$, $H'$, and $W'$ are the grid's number of channels, height, and width, which is the same in the input and output. The pointwise transformation is computed as:

\begin{equation}
 \begin{pmatrix} x_i^s \\ y_i^s  \end{pmatrix} = 
  \theta(G_i) = \theta \begin{pmatrix} x_i^t \\ y_i^t \\ 1 \end{pmatrix}
\end{equation}

where $(x_i^s, y_i^s)$ are the source coordinates in the input,  $(x_i^t, y_i^t)$ are target coordinates of the grid in the output feature map, and $\theta$ is given in equation 3. To preform the spatial transformation, the input and the sampling grid ($\theta(G)$) are sent to a bilinear sampler to produce the output map $V \in \mathbb{R}^{C \times H' \times W'}$. Since the bilinear sampling is differentiable, it allows the gradients loss to flow back to the sampling grid coordinates, and therefore to the transformation parameters $\theta$ and the localisation network ($L$). This process allows the localization of relevant ROIs in the input image. Note that although this work focuses on five ROIs ($N=5$), any region can be easily added or removed by adjusting $\theta$.

\subsection{ROIs Segmentation}
We use three parallel pathways: handcrafted pathway ($HP$), spatial (detail) pathway ($SP$), and global pathway ($GP$). 

\subsubsection{Handcrafted Pathway ($HP$) and Spatial Pathway ($SP$)}
Depending on the medical imaging modality, the regular convolutional layers can be replaced by handcrafted encoded convolutional layers. These handcrafted encoded layers can extract a unique set of statistical, geometrical, or textural features. In this work, we propose to replace the regular convolutional layers with local binary encoded (LBP) convolutional layers. LBP is an effective texture descriptor for summarizing the image's local spatial structure \cite{ojala2000gray}. The traditional LBP achieved excellent performance for texture segmentation and analysis in different medical imaging modality (e.g., \cite{iakovidis2008fuzzy,pereira2020dermoscopic}) due to its robustness to illumination changes and ability to differentiate small differences in texture and topography \cite{ojala2000gray}. The performance of LBP relies heavily on different parameters including the base, pivot, and order.  In this work, we propose to integrate LBP kernels to extract rich set of textural features and learn the optimal set of LBP parameters. Specifically, we utilized LBP encoded convolutional layers \cite{juefei2017local} to obtain unique textural features at different levels of abstraction.




Similar to \cite{juefei2017local}, the LBP encoded convolutional layer consists of a set of $m$ predefined fixed LBP filters that are used to generate $m$ difference maps. These maps are then activated using a non-linear activation (i.e., ReLU) function to generate the bitmaps. Finally, these $m$ bitmaps are combined linearly by $m$ learnable weights. As demonstrated in \cite{juefei2017local}, the number of learnable parameters in LBP encoded convolutional layer is significantly smaller (up to 169 times smaller) than the regular convolutional layer. As shown in Figure 1, $HP$ has three LBP encoded layers. Each layer includes a LBP convolution with a kernel size of 3 and a stride of 1. \textit{We hypothesize that $HP$ pathway would capture unique textural characteristics of the medical images and lead to better overall performance.}  In parallel to $HP$, $SP$ is designed to capture rich low-level details from the images at different levels of abstraction. This pathway contains three convolutional blocks, each containing a $3 \times 3$ convolutional layer with stride of $2$ followed by batch normalization and ReLU activation. The number of filters in the first, second, and third blocks are $64$, $64$, and $128$, respectively.

\subsubsection{Global Pathway ($GP$)}
As the receptive field has a great impact on segmentation, we used a lightweight model (i.e., $Xception$) for fast-downsampling of the feature map to obtain a sufficient receptive field and encode high level context information. Then, we attached a global average pooling on the tail of the lightweight model to provide the maximum receptive field with
global context information. Finally, the output of the global average pooling is up-sampled as shown in Figure 1.

\subsubsection{Fusion Module}
The feature representations of the pathways are complementary and encode different information; i.e., $HP$ (textural features), $SP$ (rich low-level details), and $GP$ (high-level context). To combine these features, we concatenate the output features from the pathways and then use batch normalization to balance their different scales. We then pooled the concatenated features into a single feature vector.

\subsection{Training}
We trained TaNet in two stages: pre-training and fine-tuning. 

In the first stage, we pre-trained the coarse segmentation and the localization network ($L$) as follows. First, we pre-trained the coarse segmentation module (FCN-8) with a batch size equal to $32$, a learning rate of $1 \times 10^{-3}$ that is reduced when the loss plateau. We used Adam to minimize the CrossEntropy loss between GT masks and the predicted coarse segmentation. Then, we used the output of the coarse segmentation as inputs to the localization network ($L$). The localization ($L$) is trained with a batch size equal to $32$ and $1 \times 10^{-3}$ learning rate. $L$ aims to optimize the loss (Smooth L1) between $\theta$ and $\bar{\theta}$ (GT). In the second stage, we loaded the pre-trained parameters from the previous stage and fine-tuned the entire network. The entire network is trained for $100$ epochs with a batch size equal to $32$. We used $1 \times 10^{-3}$ learning rate and Adam optimizer to minimize the following loss function: $L_{TaNet} = \frac{1}{N} \sum_{i}^{N} L_{SM}(\tilde{z_r}, \hat{J_r})$, where $L_{SM}$ is the softmax loss, $\tilde{z_r}$ represents the pixel-wise prediction for $ROI_r$, $r \in \{1,2,..,N\}$, $\hat{J_r}$ represents the corresponding ground truth, and $N$ represents the total number of regions.









 \begin{figure}[!t]
        \centering
        \includegraphics[height=0.174\textwidth]{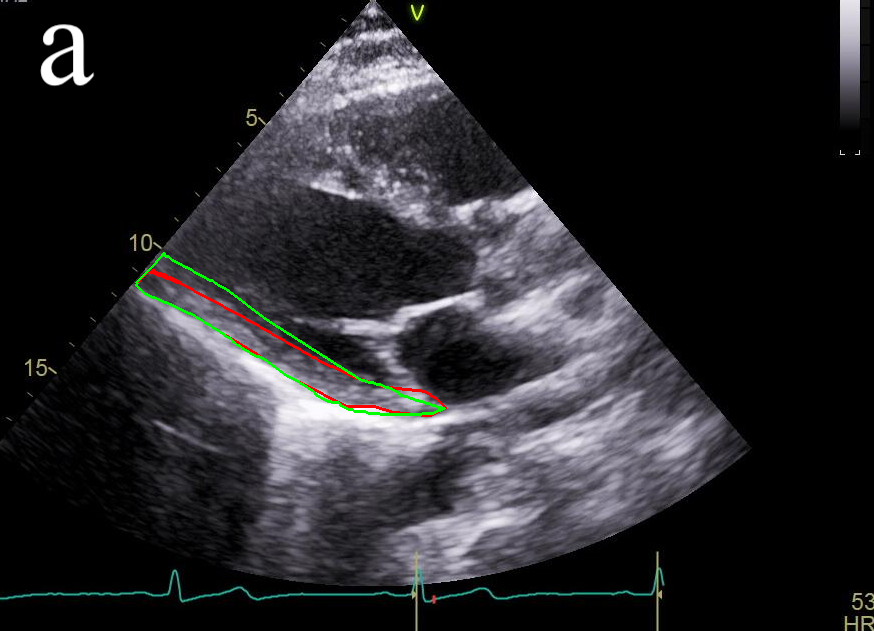}
         \includegraphics[height=0.174\textwidth]{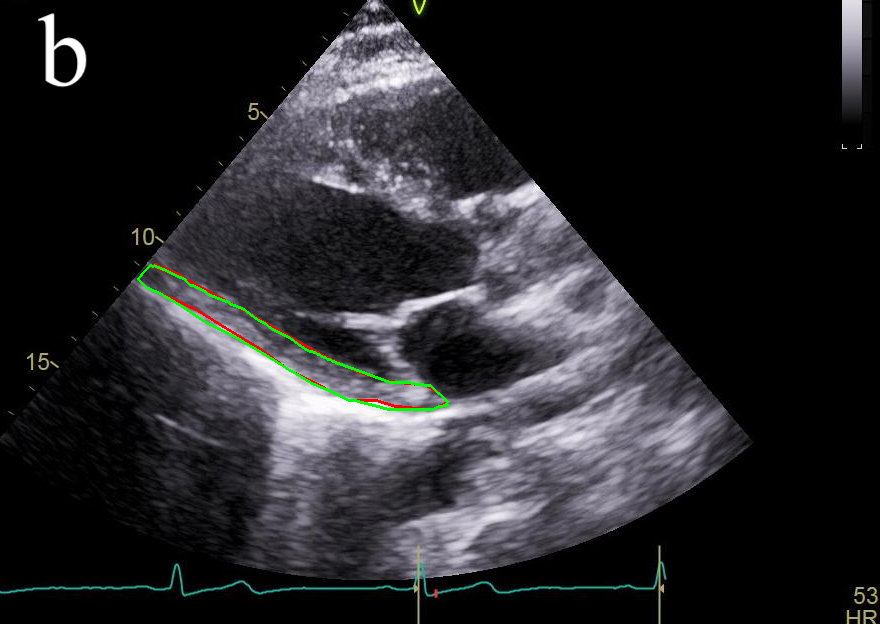}
           \includegraphics[height=0.174\textwidth]{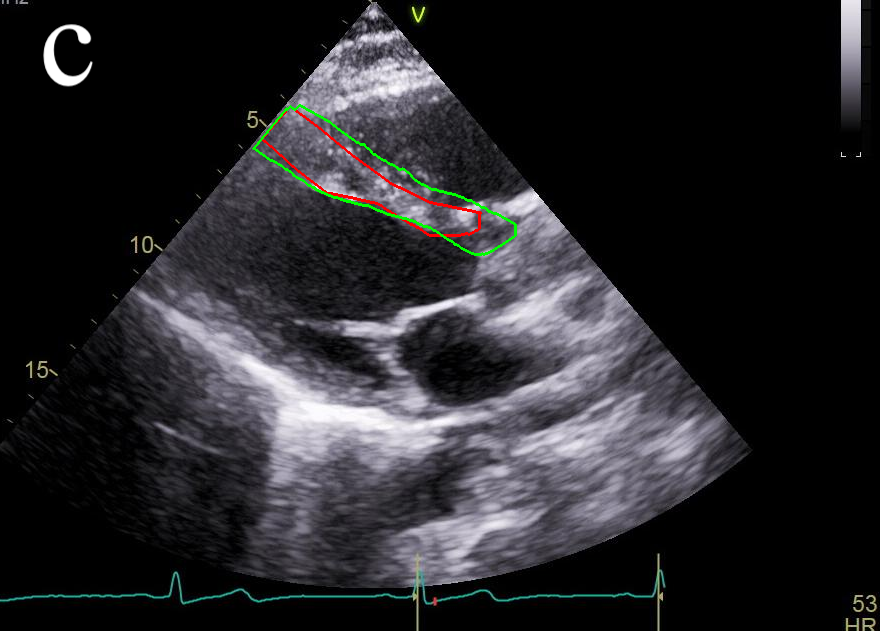}
           \includegraphics[height=0.174\textwidth]{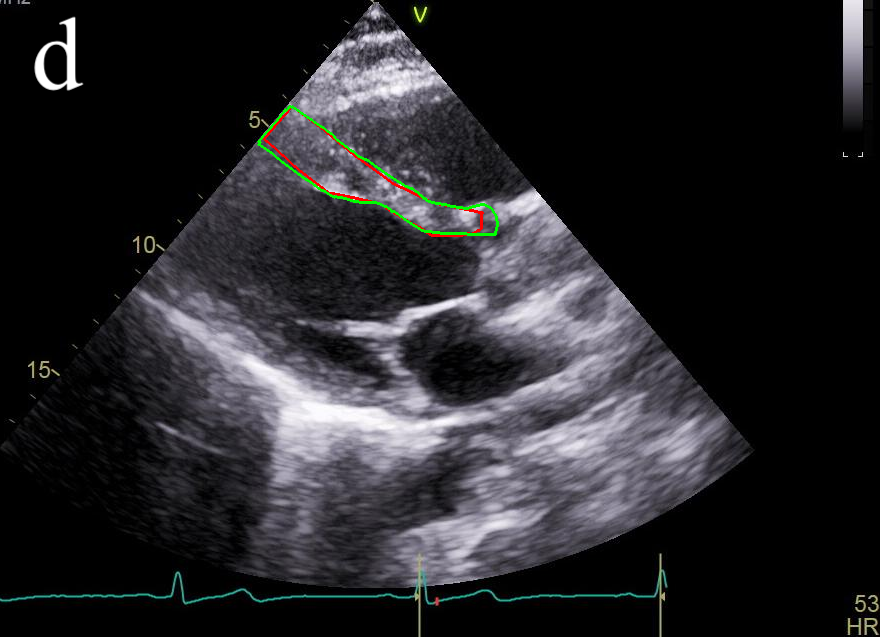}
        \caption{Segmentation of posterior and septal walls using BiSeNet and TaNet. a) PW using BiSeNet, b) PW using TaNet, c) IVS using BiSeNet, and d) IVS using TaNet. Red and green represent the ground truth and automated contours, respectively. }
        \label{fig:my_label}
    \end{figure}
    
\section{Experiments}
We conducted all experiments using Pytorch and performed training and inference on NVIDIA GTX1080Ti GPU. In all experiments, we reported the performance using mean IoU and F1, which are averaged over the test samples, and measure the speed as frames per second (FPS). Talos \cite{Talos} is used for selecting the networks (e.g., filter size) and training (e.g., batch) hyperparameters for all models. We performed ablation experiments to validate the effectiveness of STN and HP, and compared TaNet to the state-of-the-arts.

To evaluate STN impacts on the segmentation performance, we integrated STN module to Bisenet \cite{yu2018bisenet}, FCN-8 \cite{long2015fully}, and UNET \cite{ronneberger2015u} and reported the results in Table 1. The second rows of all models (i.e., FCN-8, UNET, BiSeNet) show that integrating STN slightly decreases the inference speed. However, it improves cardiac region segmentation in most cases. The red cells indicate that this improvement is significant ($p < 0.05$,  t-test), and hence, it supports our hypothesis. Specifically, these results demonstrate STN ability to increase the performance by focusing the segmentation attention on specific ROI while learning the context relationship of different ROIs. We also conducted another ablation study to evaluate the impact of LBP-encoded kernels on segmentation and reported the results in Table 1. The third rows of all models (i.e., FCN-8, UNET, BiSeNet) show that combining the textural features leads to an improvement in inference speed and performance. The blue cell indicate that the LBP-encoded kernels significantly ($ p < 0.05$,  t-test) improves the segmentation, especially in case of IVS and PW regions (see Figure 2). Due to the similarity between the anterior and posterior walls, the accurate segmentation of LV walls is challenging. However, combining the texture kernels with low-details features improves the posterior and septal walls segmentation as shown in the table and Figure 2. These results support our hypothesis and demonstrate the benefits of using $HP$ in the segmentation models. 

The last row of Table 1 presents the performance of TaNet (STN, HP, SP, GP) as compared to the baseline BiSeNet ($SP$ + $GP$) \cite{yu2018bisenet}, FCN-8 \cite{long2015fully} and UNET \cite{ronneberger2015u}. The green cells indicate that the performance of the proposed TaNet network is significantly ($p < 0.05$, t-test) higher than all baseline models. Note that TaNet has a slightly lower speed as compared to BiSeNet \cite{yu2018bisenet} due to the integration of STN module and $HP$ pathway. However, this speed is still efficient for real-time medical image analysis. We also compared TaNet with recent cardiac segmentation works: EchoNet \cite{ouyang2020video} (LV) and Res-U \cite{ali2021echocardiographic} (LV \& LA). TaNet outperformed both \cite{ouyang2020video,ali2021echocardiographic} in LV segmentation; i.e., TaNet IoU= 0.95; EchoNet IoU=0.87; Res-U IoU= 0.91. Also, TaNet outperformed \cite{ali2021echocardiographic} in LA segmentation; i.e., TaNet IoU= 0.93; Res-U IoU= 0.88. 

These results demonstrate the ability of TaNet to provide accurate medical image segmentation in real time. The conceptual design of TaNet is suitable for echo analysis as it allows to perform independent analysis for each cardiac region as well as all regions together. 


\begin{table*}[!t]
\setlength{\tabcolsep}{2.5pt}
\centering
\caption{Results of ablation experiments to evaluate the impact of STN and LBP-encoded kernels on segmentation. Red cells indicate statistical difference between models w/ and w/o STN, blue cells indicate statistical difference between models w/ and w/o LBP-encoded kernels, and green cell indicate statistical difference between TaNet and baseline models, w/o STN (\ding{53}) and LBP (\ding{53}). }\label{tab1}
\begin{tabular}{c|cc|ccccccccccc}
\hline
Model & STN & LBP & \multicolumn{2}{c}{LV} & \multicolumn{2}{c}{RV}  & \multicolumn{2}{c}{LA} & \multicolumn{2}{c}{IVS} & \multicolumn{2}{c}{PW} & FPS \\
\hline
&  &  & IoU & F1 & IoU & F1 & IoU & F1  & IoU & F1 & IoU & F1  \\
\hline

 & \ding{53} & \ding{53} & 0.87  & 0.93 & 0.82 & 0.90 & 0.86 & 0.92 & 0.86 & 0.93 & 0.76 & 0.86 & 5.5  \\
FCN-8 & $\checkmark$ &  & \cellcolor{red!15}0.93 &  \cellcolor{red!15}0.96 & 0.80  &  0.89 &  \cellcolor{red!15}0.89 &   \cellcolor{red!15}0.94 & 0.86 &   0.93  &  0.76 &  0.87 & 3.8 \\

 &  & $\checkmark$  & 0.88 & 0.94 & 0.83  &  0.91 &  0.86 &   0.93 & \cellcolor{blue!15}0.90 &   \cellcolor{blue!15}0.95  &  \cellcolor{blue!15}0.80 &  \cellcolor{blue!15}0.89 & 9.3   \\

\hline
\hline


  & \ding{53} &  \ding{53} & 0.85  & 0.91 & 0.81 & 0.89 & 0.88 & 0.93 & 0.87 & 0.93   & 0.79  & 0.88  & 4.8 \\

UNET  & $\checkmark$ & &  \cellcolor{red!15}0.90 &  \cellcolor{red!15}0.95  &  \cellcolor{red!15}0.86 &  \cellcolor{red!15}0.92 & 0.91 &  \cellcolor{red!15}0.95 &  \cellcolor{red!15}0.90  & 0.94 & \cellcolor{red!15}0.81 & 0.89  & 3.1 \\

    & & $\checkmark$ &  \cellcolor{blue!15}0.87 &  \cellcolor{blue!15}0.93  & \cellcolor{blue!15}0.86 &  \cellcolor{blue!15}0.93 & \cellcolor{blue!15}\cellcolor{blue!15}0.91 &  \cellcolor{blue!15}0.95 &  \cellcolor{blue!15}0.89  & \cellcolor{blue!15}0.94 & \cellcolor{blue!15}0.83 & \cellcolor{blue!15}0.90  & 7.2 \\

\hline
\hline
& \ding{53} & \ding{53}  & 0.89  & 0.94 & 0.85 & 0.92 & 0.89 & 0.94 & 0.85  & 0.92 & 0.77 & 0.87 & 100.1 \\

BiSeNet & $\checkmark$ & & \cellcolor{red!15}0.94 & \cellcolor{red!15}0.97  &  \cellcolor{red!15}0.90 &  \cellcolor{red!15}0.94 &  \cellcolor{red!15}0.92 &   \cellcolor{red!15}0.96 &   \cellcolor{red!15}0.90 &    \cellcolor{red!15}0.95 & 0.78 & 0.87  & 90.5 \\

 & & $\checkmark$ & \cellcolor{blue!15}0.91 & \cellcolor{blue!15}0.95  &  0.86 &  0.92 &  \cellcolor{blue!15}0.93 &   \cellcolor{blue!15}0.96 &   \cellcolor{blue!15}0.90 &    \cellcolor{blue!15}0.94 & \cellcolor{blue!15}0.86 & \cellcolor{blue!15}0.92  & 146.5  \\
\hline
\hline
TaNet & $\checkmark$ & $\checkmark$ & \cellcolor{green!10}0.95 & \cellcolor{green!10}0.98  &  \cellcolor{green!10}0.91 &  \cellcolor{green!10}0.95 & \cellcolor{green!10} 0.93 &  \cellcolor{green!10}0.97 &   \cellcolor{green!10}0.93 &    \cellcolor{green!10}0.96 & \cellcolor{green!10}0.88 & \cellcolor{green!10}0.93  & 94.5  \\
\hline 
\end{tabular}
\end{table*}

\section{Conclusion}
This work presents TaNet for real-time ROI-based segmentation in medical images. TaNet achieved superior performance when evaluated on the echo dataset for cardiac region segmentation. This superiority is attributed to STN ability to focus the segmentation attention on specific ROIs while learning the context relationship between them. Further, the use of $HP$ with texture kernels along with the $SP$ allows to extract unique and rich low-level and textural features. Although we used LBP-encoded pathway due to the superiority of LBP in texture analysis, other handcrafted descriptors or kernels (e.g., Gabor) can be easily integrated to the handcrafted pathway following the same formulation.  While our empirical results are promising, we plan to further evaluate TaNet on different settings. To benefit the research community, we make our code available, and invite researchers to contribute to this effort.



%
%
%
\bibliographystyle{splncs04}
\bibliography{ref}





\end{document}


\title{Supplemental Materials: Trilateral Attention Network for Real-time Medical Image Segmentation}}
%
%
%
%
%
\maketitle              


%
%
%





\begin{figure}[!h]
        \centering
        \includegraphics[height=0.27\textwidth]{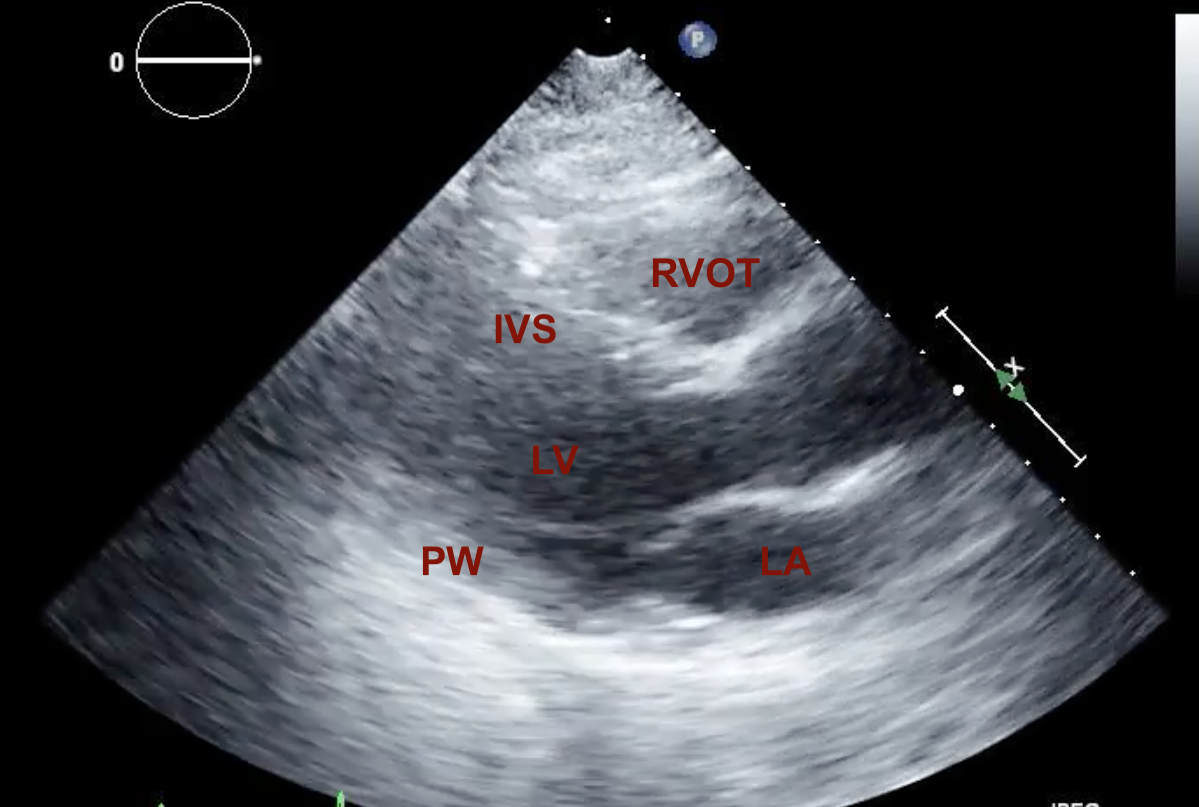}
        \includegraphics[height=0.27\textwidth]{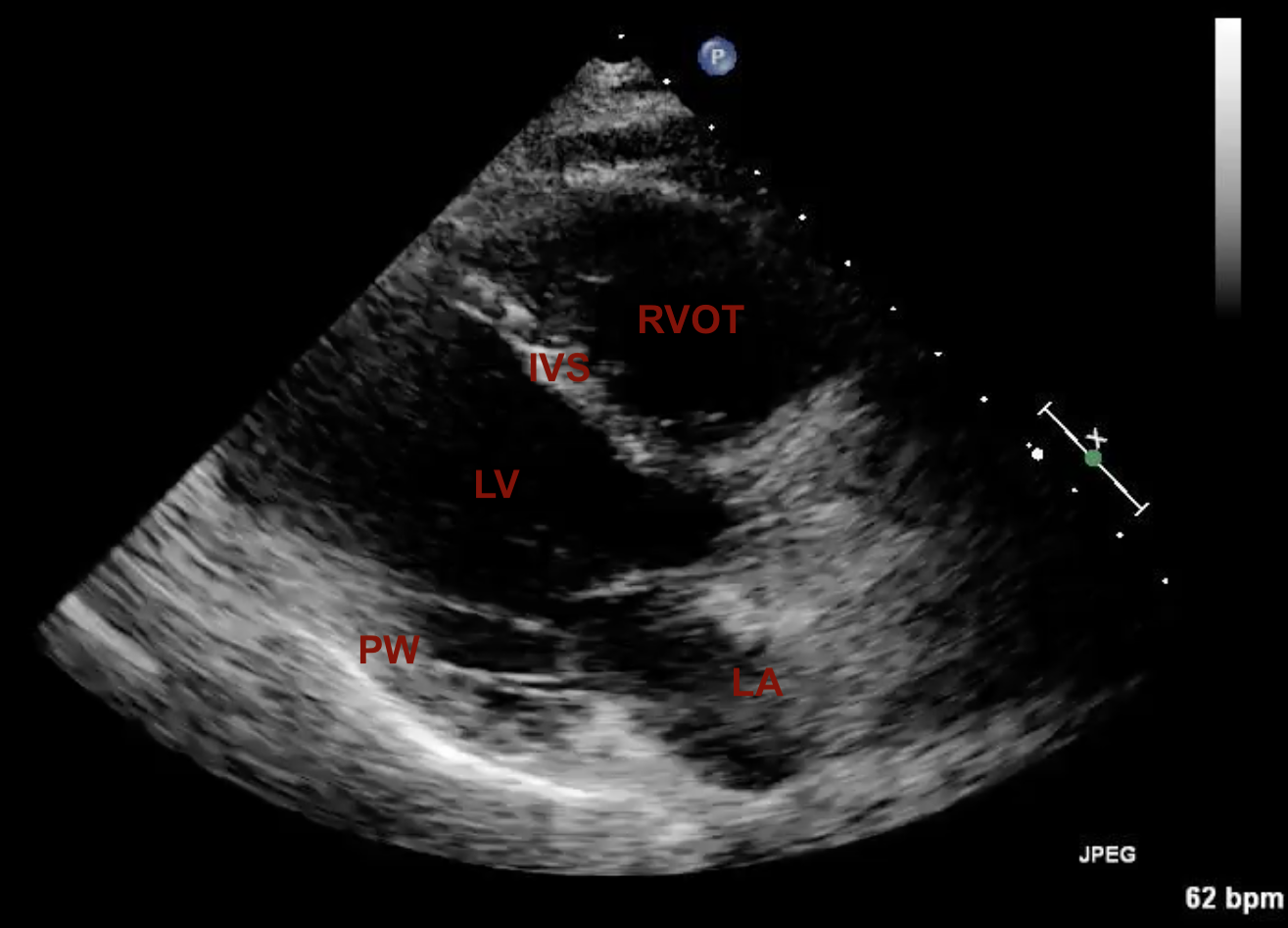}
        \caption{Examples from our dataset with low (left) and high (right) quality. LV, LA, RVOT, IVS, and PW are the left ventricle, left atrium, right ventricular outflow tract, interventricular septum, and posterior wall, respectively. }
        \label{fig:my_label}
    \end{figure}

 \begin{figure}[!b]
        \centering
        \includegraphics[height=0.15\textwidth]{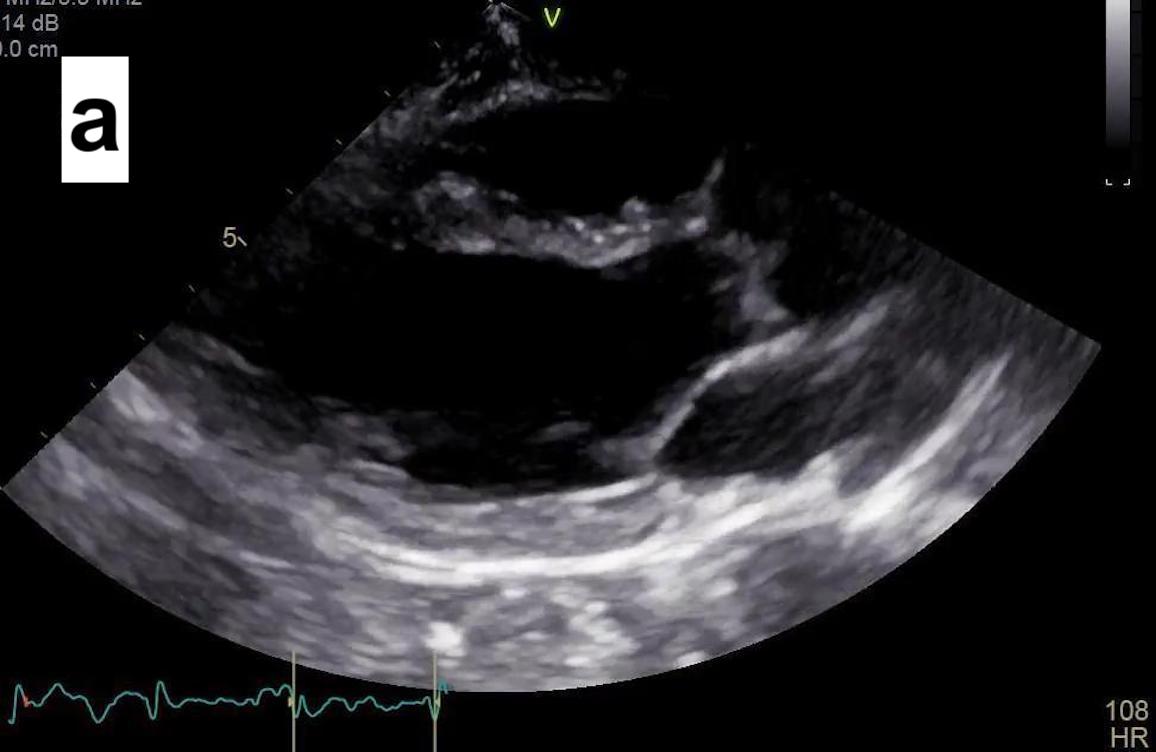}
       \includegraphics[height=0.15\textwidth]{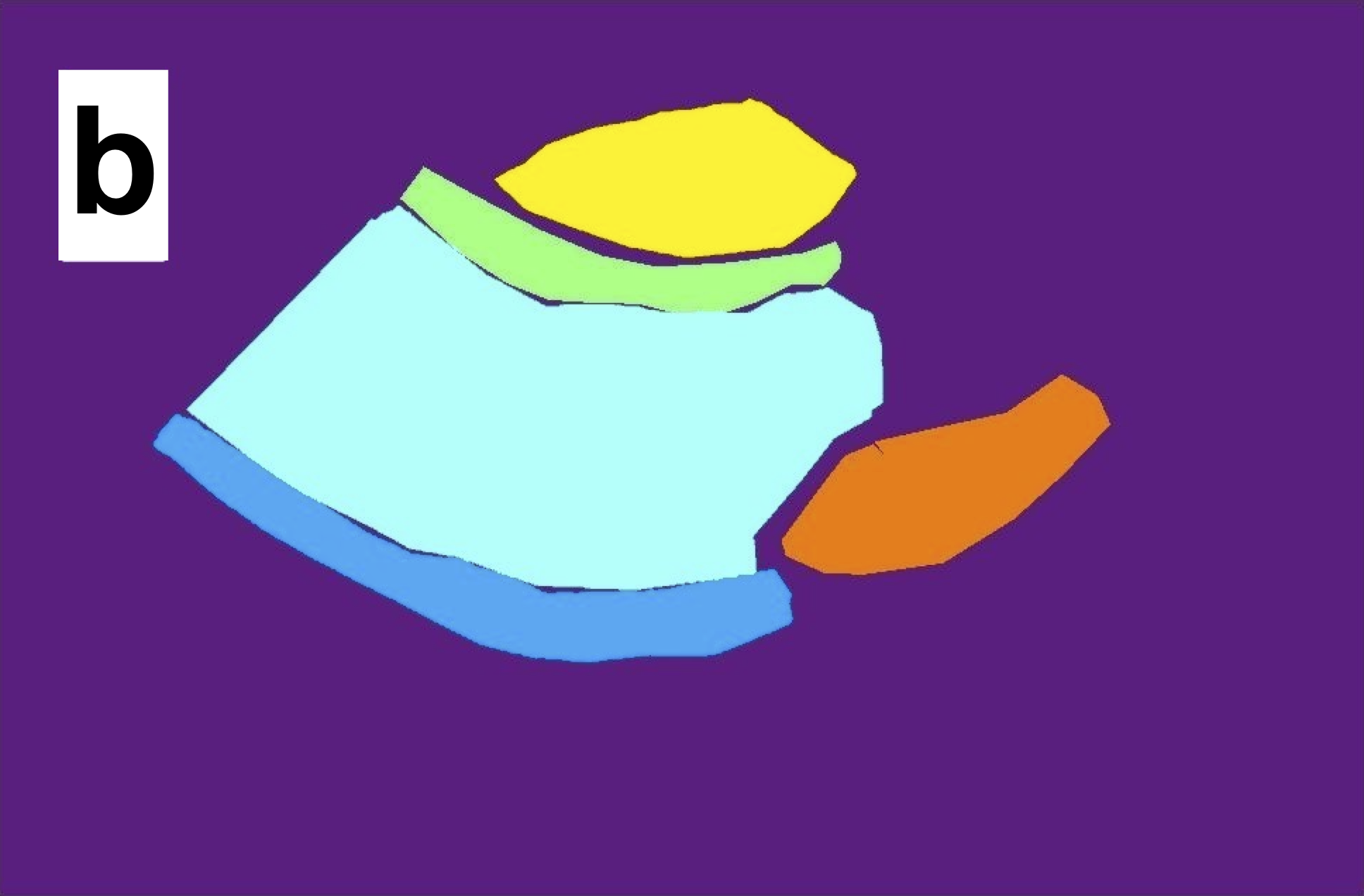}
      \includegraphics[height=0.15 \textwidth]{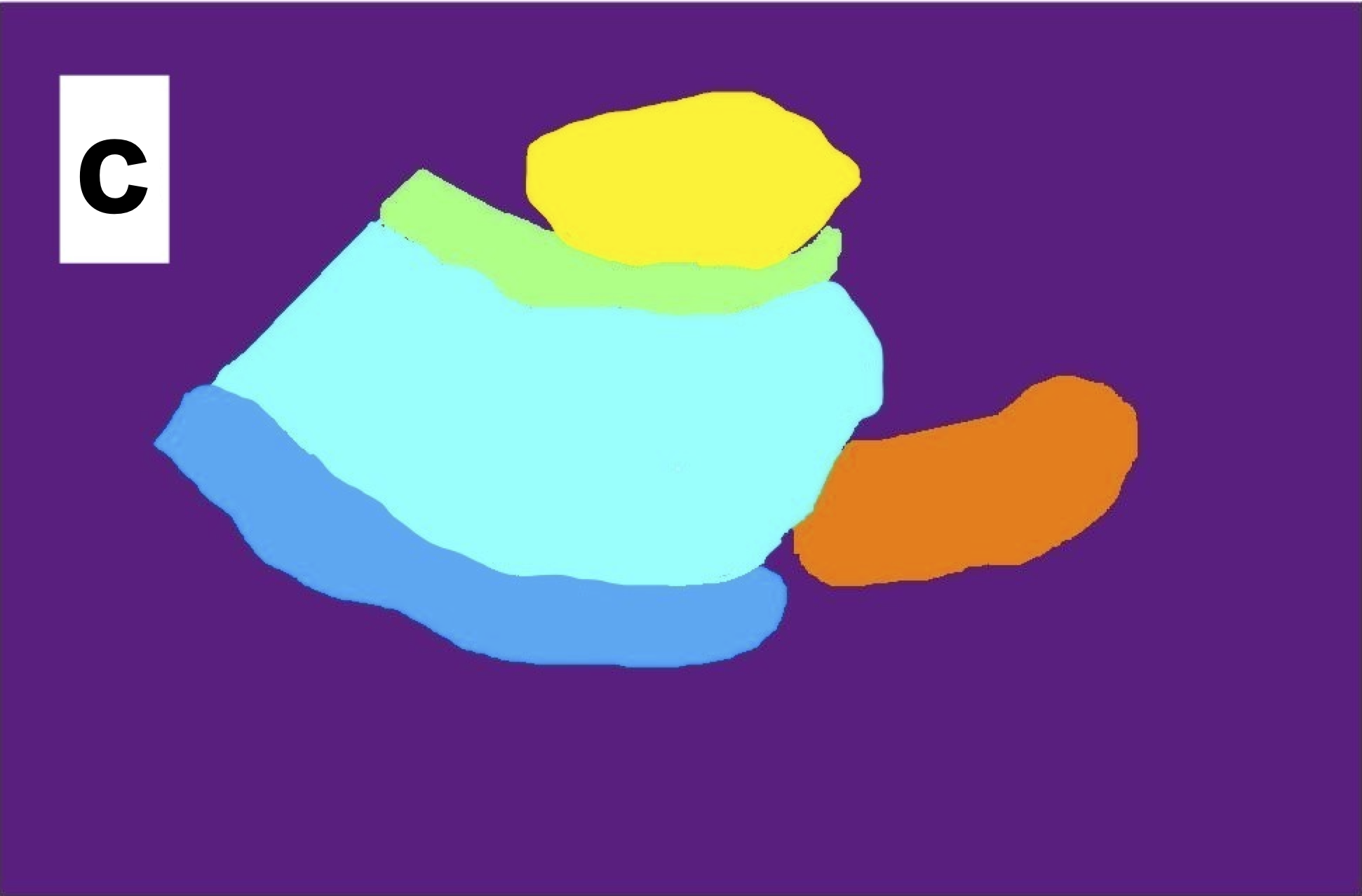}
   \includegraphics[height=0.15\textwidth]{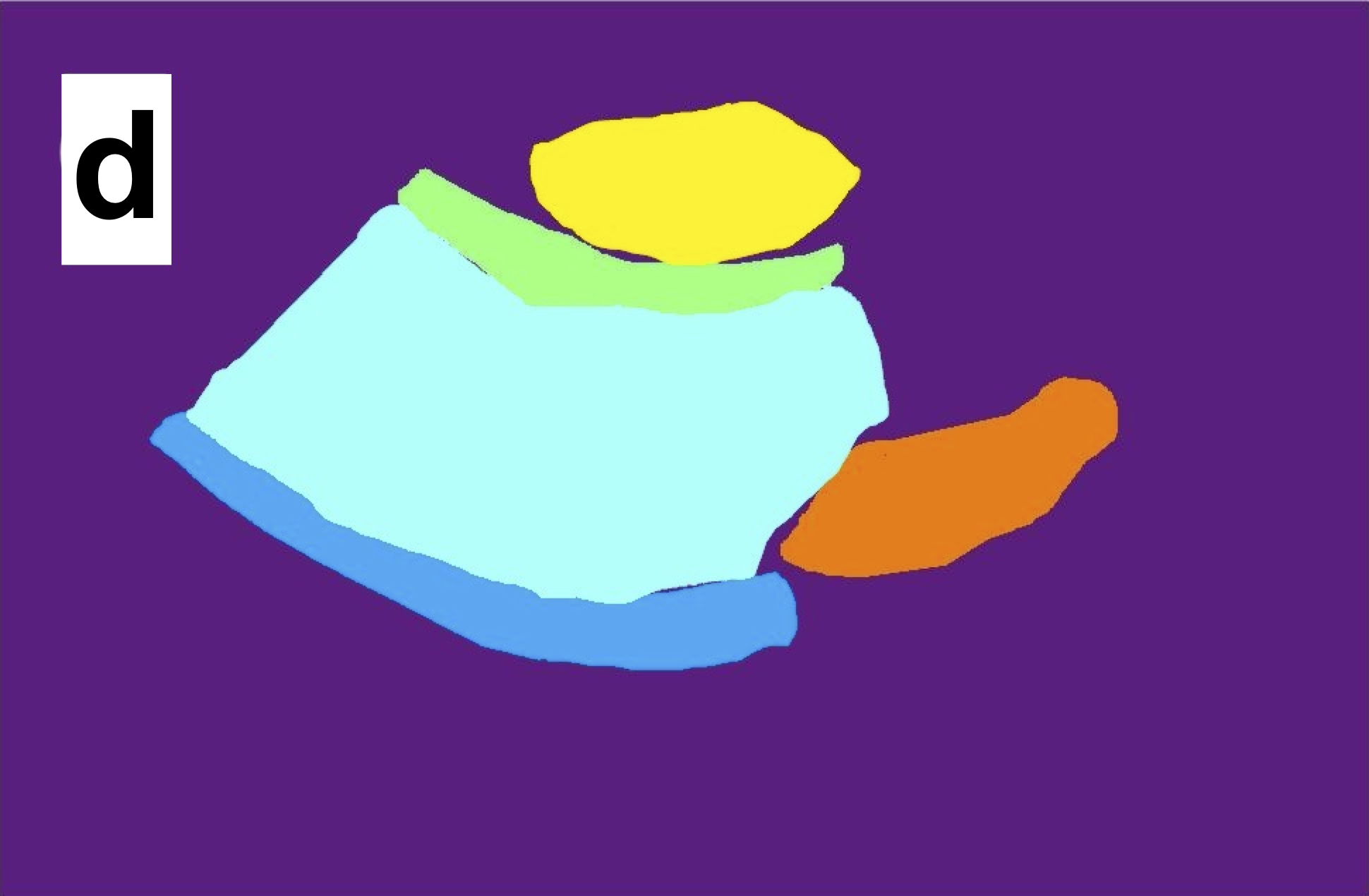}
    \caption{Examples of segmentation outputs. a) original image, b) ground truth mask, c) BiSeNet output, and d) TaNet output. As shown, BiSeNet (2.c) tends to over segment the cardiac regions and causes overlap between them as compared to the ground truth mask (2.a). For example, the LA region obtained by BiSeNet (2.c) slightly overlaps with the lower part of LV and the RV region overlaps with the IVS region. Similarly, BiSeNet segmented part of LV region (LV left side) as PW region and segmented part of IVS region as LV region (right side of LV); i.e., BiSeNet caused segmentation overlaps between LV and the surrounding walls. As discussed in the paper, the similarity between the anterior and posterior walls makes the segmentation of wall regions challenging. TaNet (2.d) provided segmentation labels that are more similar to the ground truth labels (2.a). This is attributed to TaNet's ability to focus on specific regions while using texture kernels to capture slight differences between the regions. We believe this conceptual design of TaNet is suitable for echo analysis as it allows independent analysis for each cardiac region separately as well as all regions together. More results of TaNet can be found in \href{https://github.com/TaNet}{TaNet Github page}}
    \label{fig:my_label}
\end{figure}

\begin{figure*}[!t]
\centering

\includegraphics[scale=.25]{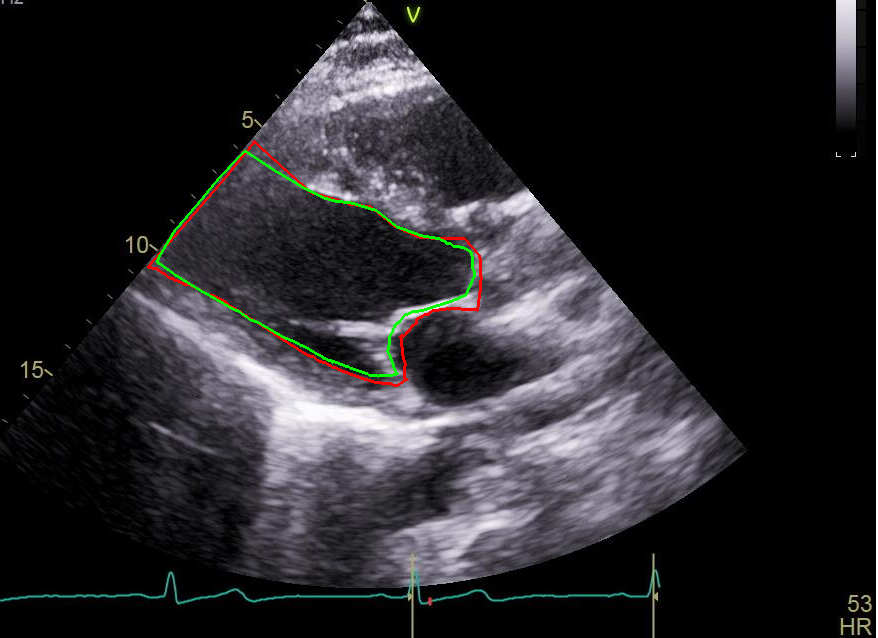}
\includegraphics[scale=.25]{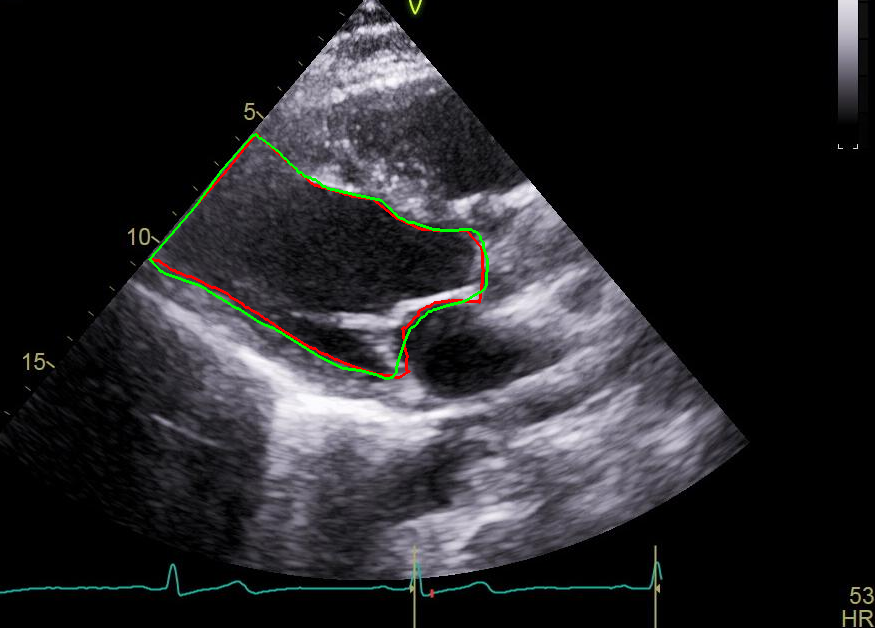}

\includegraphics[scale=.25]{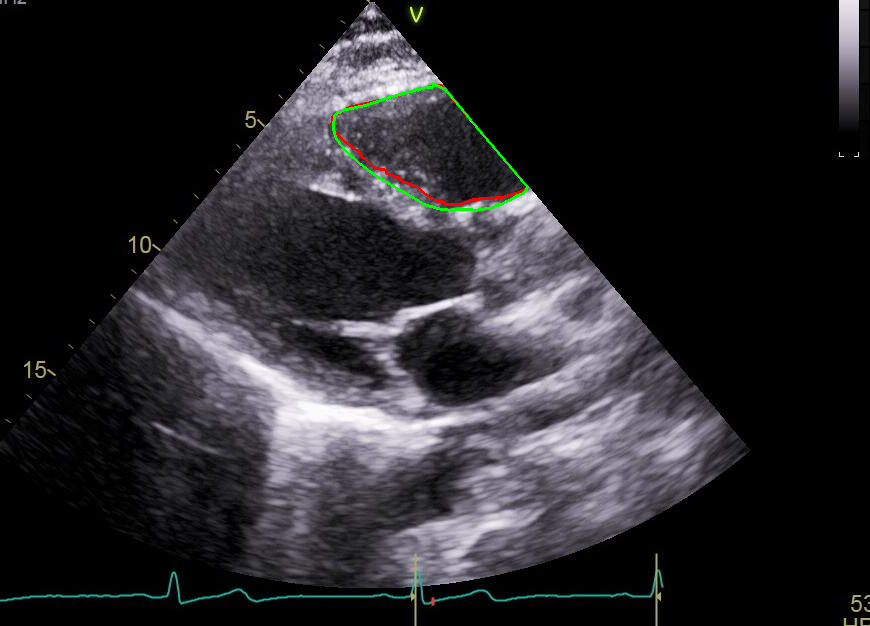}
\includegraphics[scale=.25]{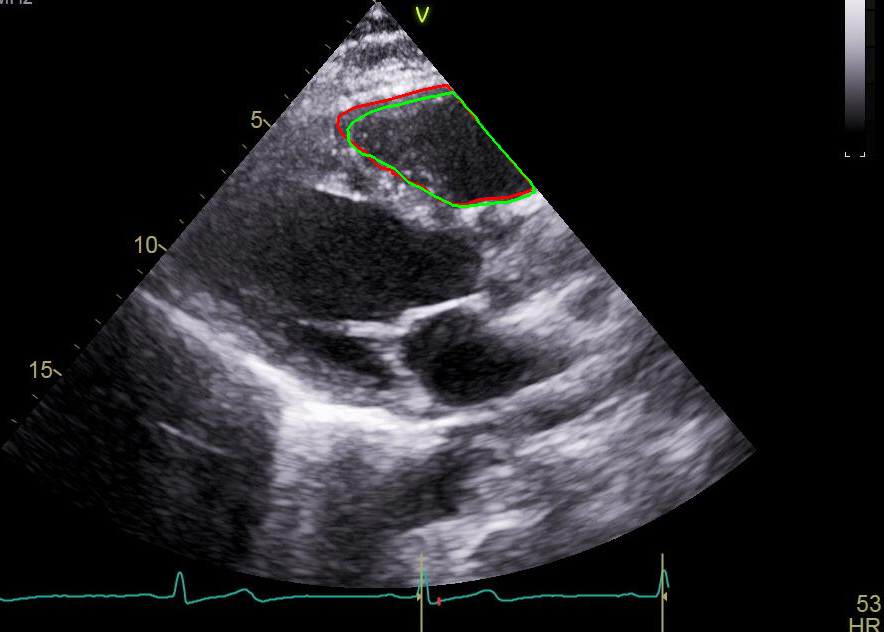}

\includegraphics[scale=.25]{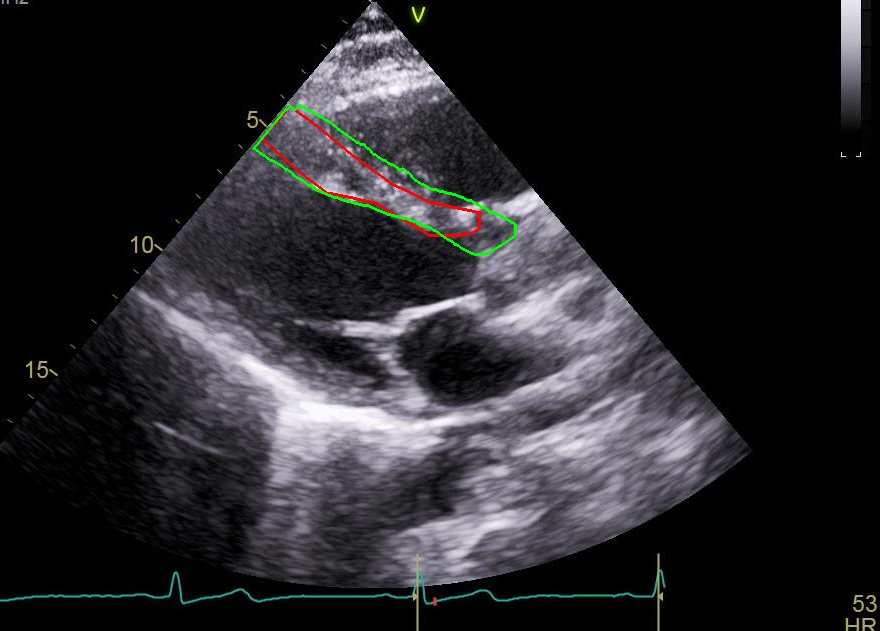}
\includegraphics[scale=.25]{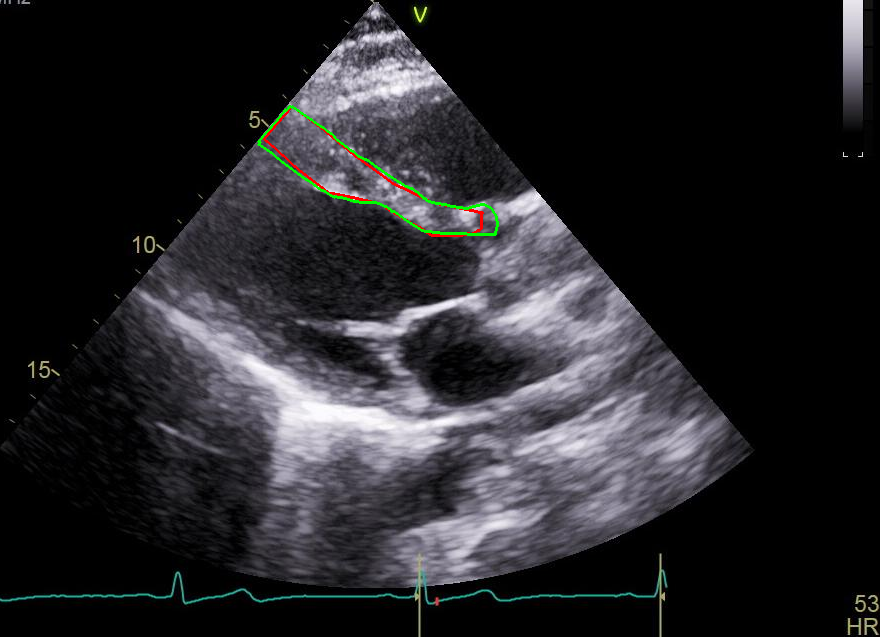}

\includegraphics[scale=.25]{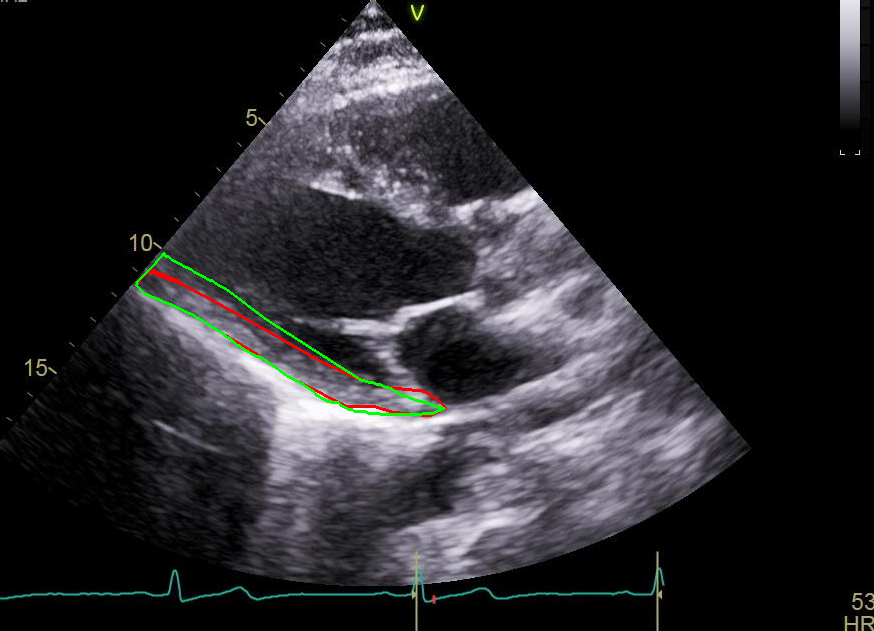}
\includegraphics[scale=.25]{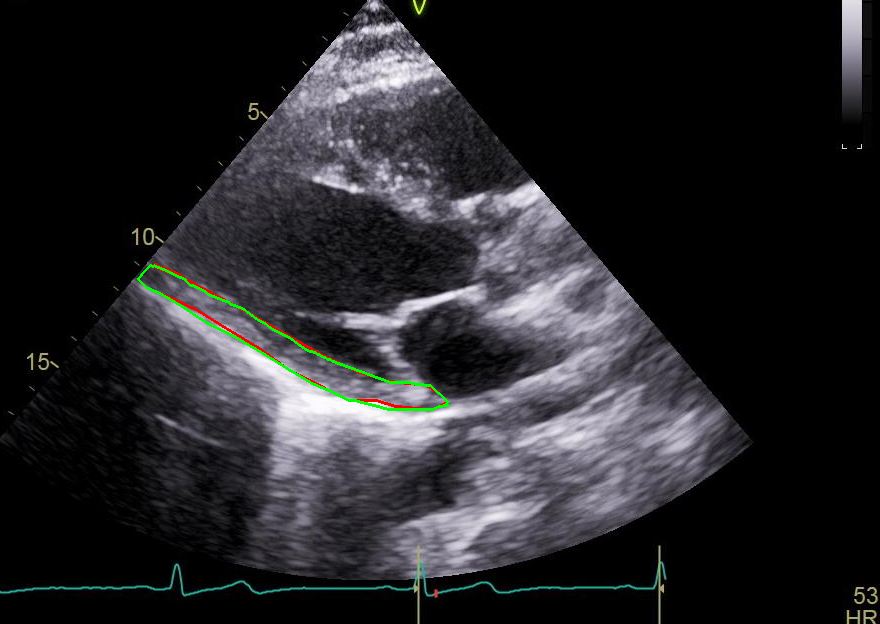}

\caption{PLAX Segmentation: Ground truth contour (red) and the contour (green) generated by BiSeNet ($1^{st}$ column) and TaNet ($2^{nd}$ column).  }
\end{figure*}